\documentclass[10pt,twocolumn,letterpaper]{article}

\usepackage[pagenumbers]{cvpr} 
\usepackage[utf8]{inputenc} 
\usepackage[T1]{fontenc}    
\usepackage{url}            
\usepackage{booktabs}       
\usepackage{amsfonts}       
\usepackage{nicefrac}       
\usepackage{microtype}      
\usepackage{xcolor}         
\usepackage{pifont}
\usepackage{graphicx}
\usepackage{amsmath}
\usepackage{amssymb}
\usepackage{adjustbox}
\usepackage{tcolorbox}
\usepackage{xspace}
\newcommand{\blue}[1]{\textcolor{blue}{#1}}
\tcbuselibrary{skins}
\usepackage{wrapfig}
\usepackage{multirow}
\usepackage{listings}
\usepackage{subcaption}  
\usepackage{mathtools}
\usepackage{soul}
\usepackage{xparse}
\usepackage{cuted}
\usepackage{algorithmic}
\usepackage{algorithm}
\definecolor{citecolor}{HTML}{0071bc}
\usepackage{tabularx}  
\usepackage{subcaption}  
\usepackage{paralist}
\usepackage{makecell}
\usepackage{xcolor,colortbl}
\definecolor{gold}{RGB}{255, 215, 0}
\usepackage{threeparttable}
\usepackage{placeins}

\newcommand{\VarSty}[1]{\textnormal{\ttfamily\color{blue!90!black}#1}\unskip}

\newcommand{\var}[1]{\texttt{#1}}
\newcommand{\For}[1]{for #1}
\newcommand{\cmarkg}{\textcolor{black}{\ding{51}}\xspace}%
\newcommand{\xmarkg}{\textcolor{gray}{\ding{55}}\xspace}%
\definecolor{myred}{HTML}{fdcdac}
\definecolor{mygreen}{HTML}{e6f5c9}
\definecolor{myyellow}{HTML}{fff2ae}

\definecolor{cvprblue}{rgb}{0.21,0.49,0.74}
\usepackage[pagebackref,breaklinks,colorlinks,allcolors=cvprblue]{hyperref}

\def\paperID{5305} 
\def\confName{CVPR}
\def\confYear{2026}

\title{RefDrone: A Challenging Benchmark for Referring Expression Comprehension in Drone Scenes}

\author{Zhichao Sun$^{1}$, Yepeng Liu$^{1}$, Zhiling Su$^{1}$, Huachao Zhu$^{1}$, Yuliang Gu$^{1}$, Yuda Zou$^{1}$, \\
Zelong Liu$^{1}$, Gui-Song Xia$^1$, Bo Du$^1$, Yongchao Xu$^1$\thanks{Corresponding author: {\tt\small yongchao.xu@whu.edu.cn}}\\
 $^1$School of Computer Science, Wuhan University
}

\begin{document}

\maketitle

\begin{abstract}
Drones have become prevalent robotic platforms with significant potential in Embodied AI. A crucial capability for drone-based Embodied AI is Referring Expression Comprehension (REC), which enables locating objects with language expressions. Despite advances in REC for ground-level scenes, drones' unique capability for broad observation introduces distinct challenges: multiple potential targets, small-scale objects, and complex environmental contexts. To address these challenges, we introduce RefDrone, an REC benchmark for drone scenes. RefDrone reveals three key challenges: 1) multi-target and no-target scenarios; 2) multi-scale and small-scale target detection; 3) complex environments with rich contextual reasoning. To efficiently construct this dataset, we develop RDAnnotator, a semi-automated annotation framework where specialized modules and human annotators collaborate through feedback loops.
RDAnnotator ensures high-quality contextual expressions while reducing annotation costs. Furthermore, we propose Number GroundingDINO (NGDINO), a novel method to handle multi-target and no-target cases. NGDINO explicitly estimates the number of objects referred to in the expression and incorporates this numerical pattern into the detection process. Comprehensive experiments with state-of-the-art REC methods demonstrate that NGDINO achieves superior performance on RefDrone, as well as on the general-domain gRefCOCO and remote sensing RSVG benchmarks.
\end{abstract}

\section{Introduction}
\label{sec:intro}

\begin{figure}
    \centering
    \includegraphics[width=1\linewidth]{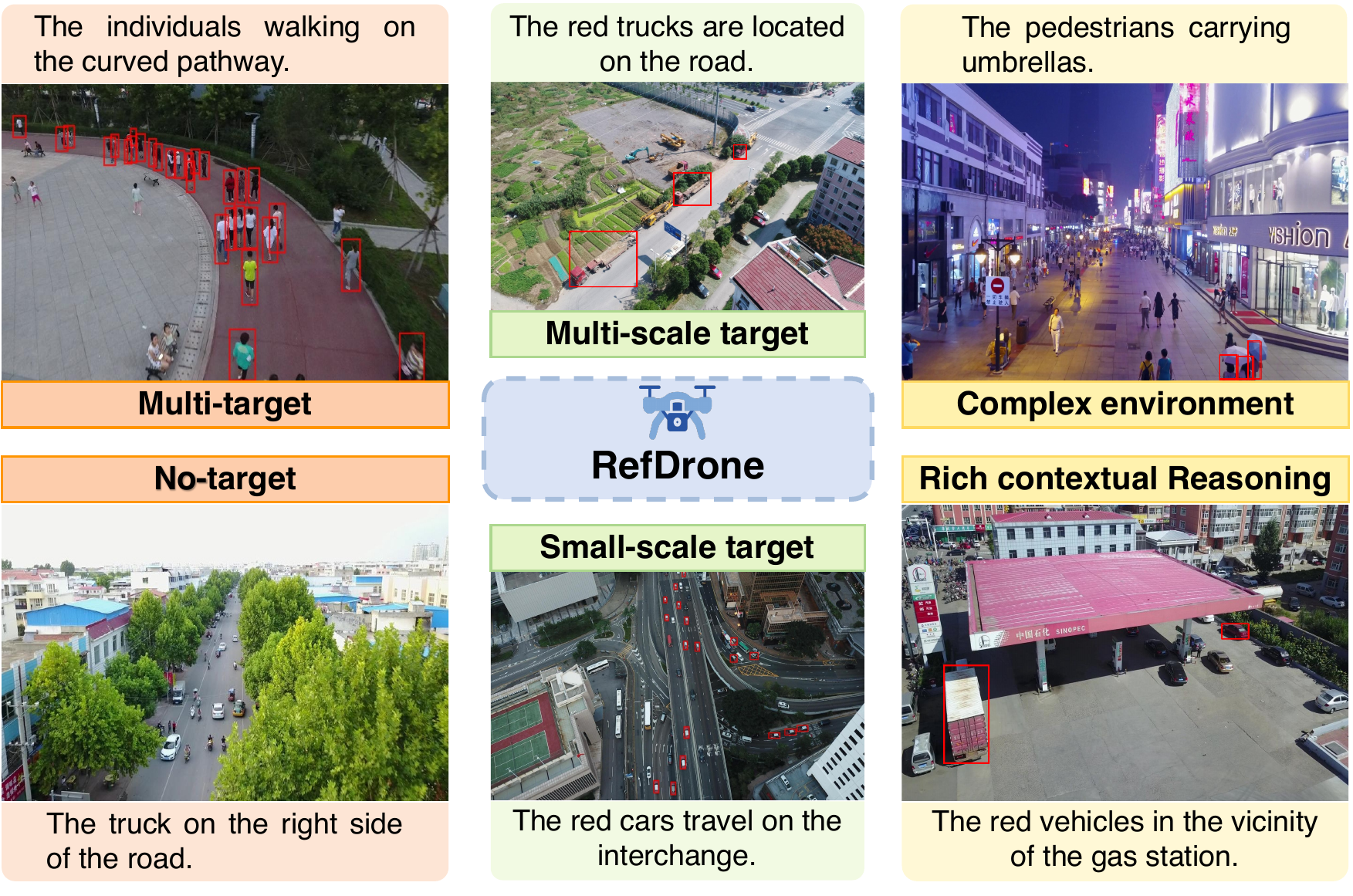}
\vspace{-15pt}
    \caption{Examples of the various challenges in RefDrone dataset.}
    \label{fig:intro-figure}
\vspace{-15pt}
\end{figure}

Drones/UAVs have become increasingly prevalent in both personal and professional applications, including entertainment, package delivery, traffic surveillance, and emergency rescue~\cite{adao2017hyperspectral, san2018uav, varga2022seadronessee}. Their mobility and broad observational capabilities make them promising platforms for Embodied AI applications~\cite{fan-etal-2023-aerial, liu2023aerialvln, lee2024citynav, gao2024embodied, liu2024coherent}. A fundamental capability in Embodied AI is Referring Expression Comprehension (REC)~\cite{sima2023embodied, Wang_2024_CVPR, cai2024bridging, wang2024polaris}, which bridges natural language understanding and visual perception by localizing specific objects in images based on textual descriptions. However, existing REC datasets predominantly adopt ground-level perspectives (\eg, RefCOCO~\cite{refcoco}), leaving drone-based scenarios largely underexplored despite their unique challenges.\par

\begin{table*}[ht]
\caption{Comparison of REC datasets relevant to RefDrone. Avg. objects: objects per expression. Avg. length: words per expression. No target: expressions without referred objects. Expression type: expression generation method. Small target: percentage of small-scale objects.}
\vspace{-8pt}
  \centering
  \begin{adjustbox}{width=\textwidth}
  \begin{tabular}{@{}l @{\hspace{2pt}}| @{\hspace{2pt}}c@{\hspace{8pt}}c@{\hspace{8pt}}c @{\hspace{8pt}}c@{\hspace{8pt}}c@{\hspace{8pt}}c@{}}
    \toprule
              & RefCOCO/+/g~\cite{refcoco,refcocog} & gRefCOCO~\cite{grefcoco} & D$^3$~\cite{xie2024described} & RIS-CQ~\cite{riscq} & RSVG~\cite{zhan2023rsvg}  & RefDrone (Ours) \\
    \midrule
 \textbf{Image source} & COCO~\cite{coco}  & COCO & COCO  & VG~\cite{vg}+COCO & DIOR~\cite{dior}  & VisDrone~\cite{visdrone} \\
 \textbf{Avg. objects}   &   1.0     & 1.4  & 1.3   &  3.6    &   2.2  & 3.8 \\
 \textbf{Avg. length} & 3.6/3.5/8.4  & 4.9  & 6.3 & 13.2 & 7.5  & 9.0 \\
 \textbf{No target}  & \xmarkg  & \cmarkg & \cmarkg & \xmarkg  & \xmarkg    & \cmarkg \\
 \textbf{Expression type} & Manual & Manual & Manual  & LLM & Templated   & LVLM \\
 \textbf{Small target}  & 0/0/0\% & 0.1\% & 6.3\%  & - & 17.2\%  & 31.1\% \\
  
    \bottomrule
   \end{tabular} 
\end{adjustbox}
  \label{tab:dataset_ann}
\vspace{-8pt}
\end{table*}

In this work, we introduce \textbf{RefDrone}, a challenging REC benchmark for drone scenes. RefDrone comprises 17,900 referring expressions annotated across 8,536 images, with 63,679 object instances. As illustrated in Figure~\ref{fig:intro-figure}, RefDrone presents three primary challenges: (1)\textbf{multi-target and no-target samples}, where expressions may refer to any number of objects (0 to 242); (2) \textbf{multi-scale and small-scale target detection}, featuring 31\% small objects and 14\% large objects; and (3) \textbf{complex environment with rich contextual reasoning}, encompassing diverse viewpoints, lighting conditions, intricate backgrounds, and rich descriptions of spatial relations, object attributes, and inter-object interactions. As shown in Table~\ref{tab:dataset_ann}, RefDrone offers greater diversity and complexity than existing REC benchmarks. We evaluate \textbf{26 representative REC models} including: specialized REC methods, task-specific LVLMs,  general LVLMs, and closed-source API models, using zero-shot settings. All models exhibit poorer performance on RefDrone compared to standard REC datasets (\eg, Qwen2.5-VL-7B~\cite{bai2025qwen2} yields  26.52\% $Acc_{img.}$ on RefDrone vs. 92.5\% $Acc_{img.}$ on RefCOCO$_\text{testA}$), highlighting the inherent difficulty and unique challenges.

\par
To enable efficient dataset construction, we develop RDAnnotator, a semi-automated annotation pipeline for referring expression annotation in drone scenes. RDAnnotator leverages multiple specialized LVLM-based modules, which collaborate within a feedback loop to generate and validate annotations. By reducing human involvement to quality control and minor refinements, RDAnnotator achieves a cost of merely \textbf{\$0.0539 per expression} (GPT-4o API) and reduces human annotation time to under \textbf{one minute per expression}. This cost-efficiency makes RDAnnotator a scalable solution for large-scale dataset construction and can be readily adapted to other REC tasks.

\par
Furthermore, we propose Number GroundingDINO (NGDINO) to address multi-target and no-target challenges. Our key insight is that \textbf{\emph{explicitly modeling the number of referred objects significantly enhances handling of these scenarios}}. NGDINO includes three components: (1) a number prediction head estimating target object counts, (2) learnable number-queries capturing numerical patterns for varying object quantities, and (3) a number cross-attention module fusing number queries with detection queries for enhanced localization. Extensive experiments on our RefDrone benchmark, as well as the general-domain gRefCOCO~\cite{grefcoco} and remote sensing RSVG~\cite{zhan2023rsvg} datasets, demonstrate that NGDINO achieves substantial improvements, particularly in multi-target and no-target cases.\par

In summary, our contributions are listed as follows:
\begin{compactitem}

    \item \textbf{RefDrone Benchmark}: We introduce RefDrone, a comprehensive REC benchmark for drone scenes featuring three key challenges: multi-target/no-target scenarios, multi-scale/small-object detection, and complex contextual reasoning. We provide thorough evaluations of 26 representative REC models.

    \item \textbf{RDAnnotator Framework}: We propose RDAnnotator, a cost-effective semi-automated annotation framework that significantly reduces human effort while ensuring high-quality annotations. This framework is scalable for large-scale dataset construction and generalizable beyond drone imagery.
    
    \item \textbf{NGDINO Method}: We develop NGDINO, a novel approach explicitly modeling object counts to address multi-target and no-target cases. NGDINO achieves state-of-the-art performance on RefDrone with consistent improvements on gRefCOCO and RSVG.
    
\end{compactitem}

\section{Related Works}

\subsection{Referring expression understanding datasets}
Referring expression understanding identifies visual regions corresponding to natural language expressions. Subtasks include Referring Expression Comprehension (REC), which predicts bounding boxes, and Referring Expression Segmentation (RES), which generates pixel-level masks. Early benchmarks such as ReferIt~\cite{kazemzadeh2014referitgame} and RefCOCO~\cite{refcoco} were pioneering but mostly limited to single-target scenarios. Subsequent works introduced greater complexity. For instance, gRefCOCO~\cite{grefcoco} expanded to multi-target expressions, while D$^3$ ~\cite{xie2024described} and RIS-CQ~\cite{riscq} provided more descriptive and challenging language. However, these datasets typically involve a small, fixed number of targets per expression.

Domain-specific extensions have also emerged, including video action recognition (RAVAR~\cite{peng2024ravar}) and affordance detection (RIO~\cite{qu2024rio}). In the aerial domain, RSVG~\cite{zhan2023rsvg} focuses on remote sensing data but uses expressions with limited contextual richness. AerialVG~\cite{Liu_2025_ICCV} focuses on using spatial relations for a single target, leaving the challenge of multi-target and no-target tasks. To fill this gap, our RefDrone benchmark provides a more complex and realistic setting by introducing complex multi-target scenarios with contextually rich expressions. Table~\ref{tab:dataset_ann} provides a detailed comparison with related benchmarks.

 \begin{figure*}[ht]
\centering
\includegraphics[width = 0.95\linewidth]{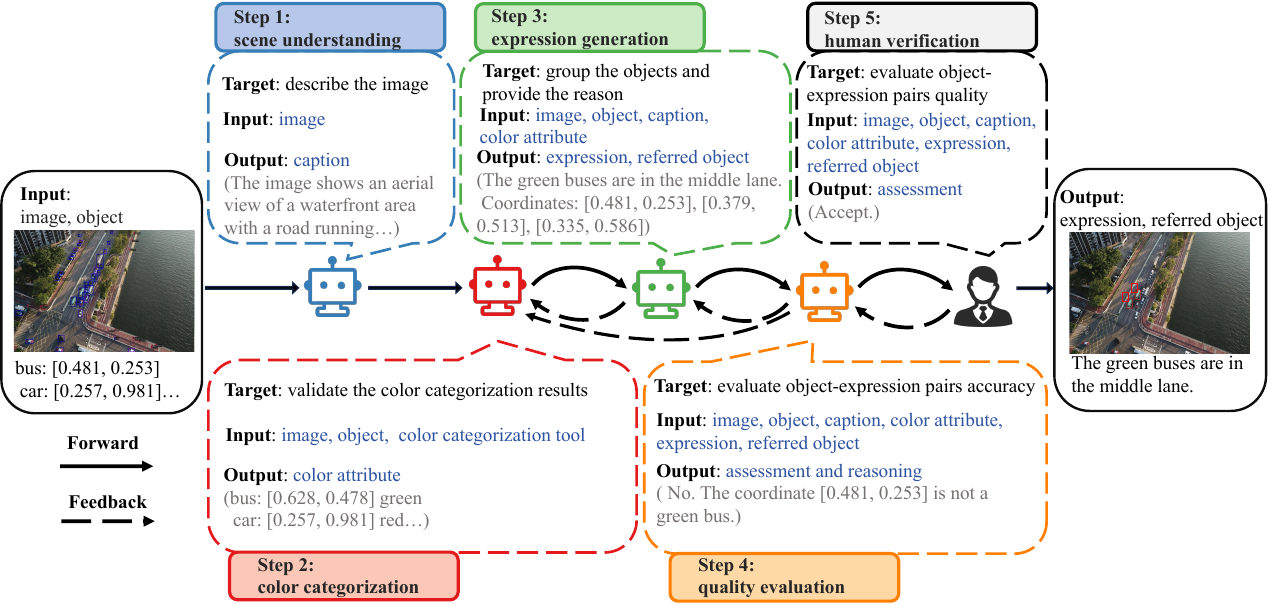}
\vspace{-5pt}
\caption{The overview of the RefDrone annotation process with RDAnnotator. Multiple specialized LVLM-based modules collaborate both with each other and human annotators through iterative feedback loops to generate high-quality annotations.}
\vspace{-15pt}
\label{fig:agent}
\end{figure*}

\subsection{Referring expression comprehension methods}
REC methods can be broadly categorized into large vision language models (LVLMs) and specialist models. 
LVLMs~\cite{you2023ferret, chen2023minigpt, chen2023shikra, bai2023qwen, peng2023kosmos, lin2023sphinx, zhan2025griffon, wang2023cogvlm, jiang2025detect} have recently been applied to REC tasks as part of evaluating their broader visual-language understanding capabilities. These models leverage extensive referring instruction tuning data to achieve competitive performance without task-specific architectural designs. To manage computational demands, LVLMs typically process downsampled images~\cite{llava} or employ visual token reduction strategies~\cite{bai2023qwen, wu2024deepseekvl2mixtureofexpertsvisionlanguagemodels}, which limits their ability to preserve fine-grained details. This constraint makes them less effective for grounding small objects, a crucial requirement in applications like aerial image analysis.

Specialist models include two-stage and one-stage methods. Two-stage methods~\cite{hu2017modeling, liu2019learning, hong2019learning, reclip, han2024zero} typically approach REC as a ranking task: first generating region proposals through an object detector, then ranking proposals based on language-vision alignment. Despite achieving strong accuracy, two-stage pipelines suffer from slow inference. In contrast, one-stage methods~\cite{mdetr, glip, liu2023dq, gan2020large, yan2023universal, gdino} directly predict target regions guided by language input. These approaches leverage transformers to enable cross-modal interactions between visual and textual features. Among these, GroundingDINO (GDINO)~\cite{gdino} has gained widespread attention for its impressive results in REC tasks. Our work extends GDINO by incorporating explicit number modeling to handle multi-target and no-target scenarios, which are critical challenges in drone-based REC tasks.\par

\section{RefDrone benchmark}

\subsection{Data source}
The RefDrone benchmark is built upon VisDrone2019-DET~\cite{visdrone}, a high-quality drone-captured object detection dataset. Images are collected across scenarios, illumination conditions, and flying altitudes. To ensure sufficient complexity for referring expression annotation, we filtered the dataset, retaining only images with at least three objects and excluding objects with bounding box areas smaller than 64 pixels.
VisDrone2019-DET provides object categories and bounding box coordinates, which we convert to normalized center points (range 0-1). This approach reduces the token count for LVLMs while preserving spatial relationships.\par

\subsection{RDAnnotator for semi-automated annotation}
To construct our benchmark, we introduce RDAnnotator, a semi-automated annotation framework that balances annotation quality and scalability by integrating LVLM modules with a human-in-the-loop validation process. As illustrated in Figure~\ref{fig:agent}, the framework operates in five steps.\par

\textbf{Step 1: scene understanding.} A scene-parsing module, powered by GPT-4o, generates three diverse captions for each image. These captions establish a foundational context by detailing spatial arrangements and object relationships for subsequent referring expression generation.  \par

\textbf{Step 2: color categorization.} To identify color attributes, which are crucial discriminative features in REC, we employ a hybrid pipeline. This pipeline combines a CNN-based classifier (WideResNet-101~\cite{wideresnet}) with high-level LVLM-based semantic reasoning to ensure accurate color attribution despite challenging lighting and atmospheric conditions.\par
   
\textbf{Step 3: expression generation.} We reformulate expression generation as an \emph{object grouping task}. This module clusters semantically related objects and generates a textual justification for each grouping, which then serves as the referring expression. A dynamic feedback loop connects this step to color categorization (Step 2), triggering re-analysis if novel color terms are generated, thereby ensuring attribute consistency across the dataset.\par

\textbf{Step 4: quality evaluation.}  A validation module automatically assesses each generated object-expression pair for semantic accuracy and uniqueness. Correct annotations advance to human verification (Step 5). Incorrect annotations are routed back with targeted feedback: semantic errors are returned to expression generation (Step 3), while color inaccuracies are sent back to color categorization (Step 2).\par

\begin{figure}[t]
\centering
\includegraphics[width = 1\linewidth]{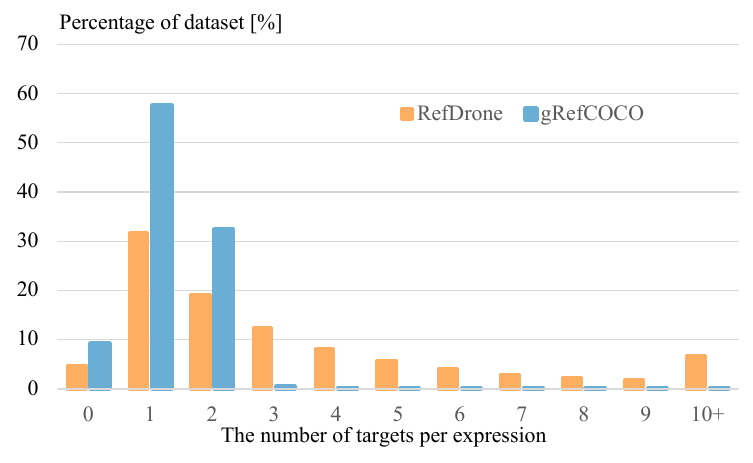}
\vspace{-20pt}
\caption{Object number distribution per expression in gRefCOCO~\cite{grefcoco} and RefDrone datasets.
} 
\vspace{-10pt}
\label{fig:data_statistics}
\end{figure}

\begin{figure}[t]
\subfloat[{\scriptsize}]{
    \begin{minipage}[l]{0.29\linewidth}
         \raggedright
             \vspace{-100pt} 
        \includegraphics[width=1\linewidth]{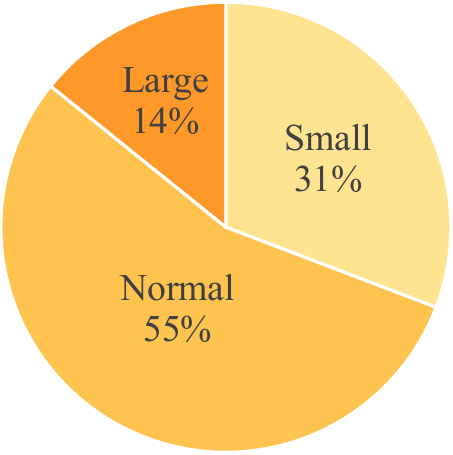}
    \end{minipage}
}
\subfloat[{\scriptsize }]{
    \begin{minipage}[l]{0.69\linewidth}
         \raggedleft
        \includegraphics[width=1.0\linewidth]{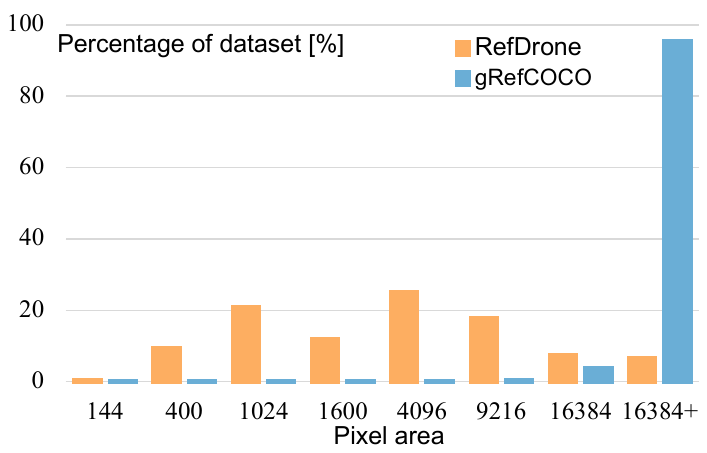}

    \end{minipage}
}
\vspace{-5pt}
    \caption{Object size distribution analysis. (a) Object size distribution in RefDrone dataset (small: $< 32^2 = 1024$ pixels, normal: 1024 to 9216 pixels, large: $> 96^2 = 9216$ pixels). (b) Object size histograms in RefDrone and gRefCOCO~\cite{grefcoco} datasets.}
    \label{fig:data_area}
\vspace{-10pt}
\end{figure}

\textbf{Step 5: human verification.} All generated annotations undergo a final human review, categorized into three tiers:
\begin{compactitem}
\item \textit{Direct acceptance}. Annotations satisfying all criteria are approved for the final dataset.
\item \textit{Refinement required}. Annotations with minor errors are corrected through human editing.
\item \textit{Significant issues}. Annotations with poorly grounded or inconsistent content trigger full regeneration.
\end{compactitem}
This multi-level feedback system ensures high annotation fidelity. Expressions that consistently fail verification are designated as ``no-target'' samples, creating a set of contextually valid negative examples.

\par

Each step leverages LVLMs via in-context learning with carefully designed task-specific prompts and examples (see Appendix). The effectiveness of our framework is demonstrated by the annotation outcomes: 42\% of samples were directly accepted, 47\% required minor refinement, and only 11\% were rejected for re-annotation. Ultimately, RDAnnotator reduces human annotation effort by 85\% , decreasing the average time per expression from 7 minutes to 1 minute. This efficiency is achieved at a low API cost of \$0.0539 per expression, demonstrating the framework's scalability for creating large-scale, high-fidelity REC datasets.



\begin{figure}[t]
    \centering
    \subfloat[{\scriptsize Word cloud of complete expression.}]{
    \begin{minipage}[c]{0.49\linewidth}
        \includegraphics[width=1.0\linewidth]{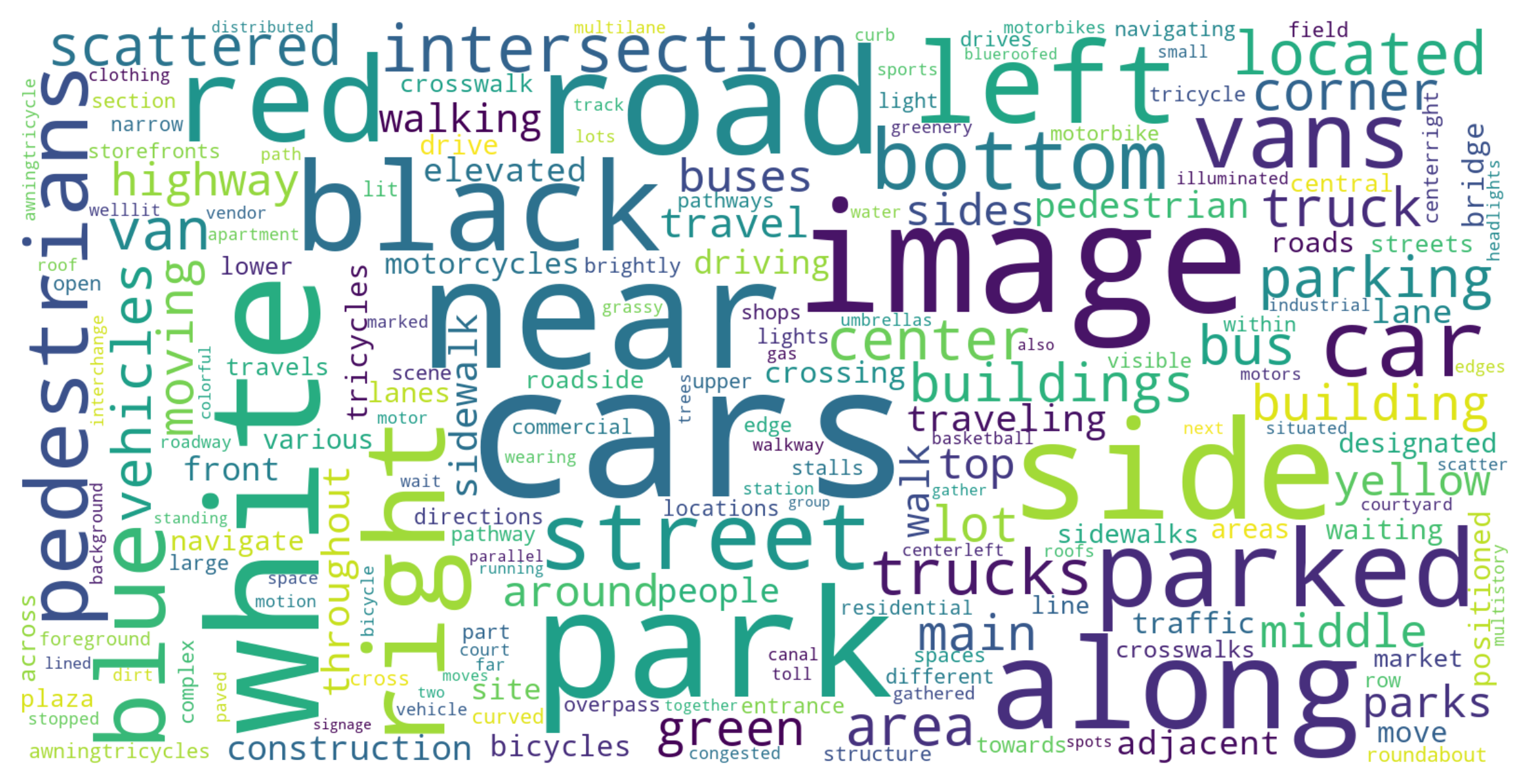}
    \end{minipage}
    }
    \subfloat[{\scriptsize Word cloud of background terms.}]{
    \begin{minipage}[c]{0.49\linewidth}

        \includegraphics[width=1.0\linewidth]{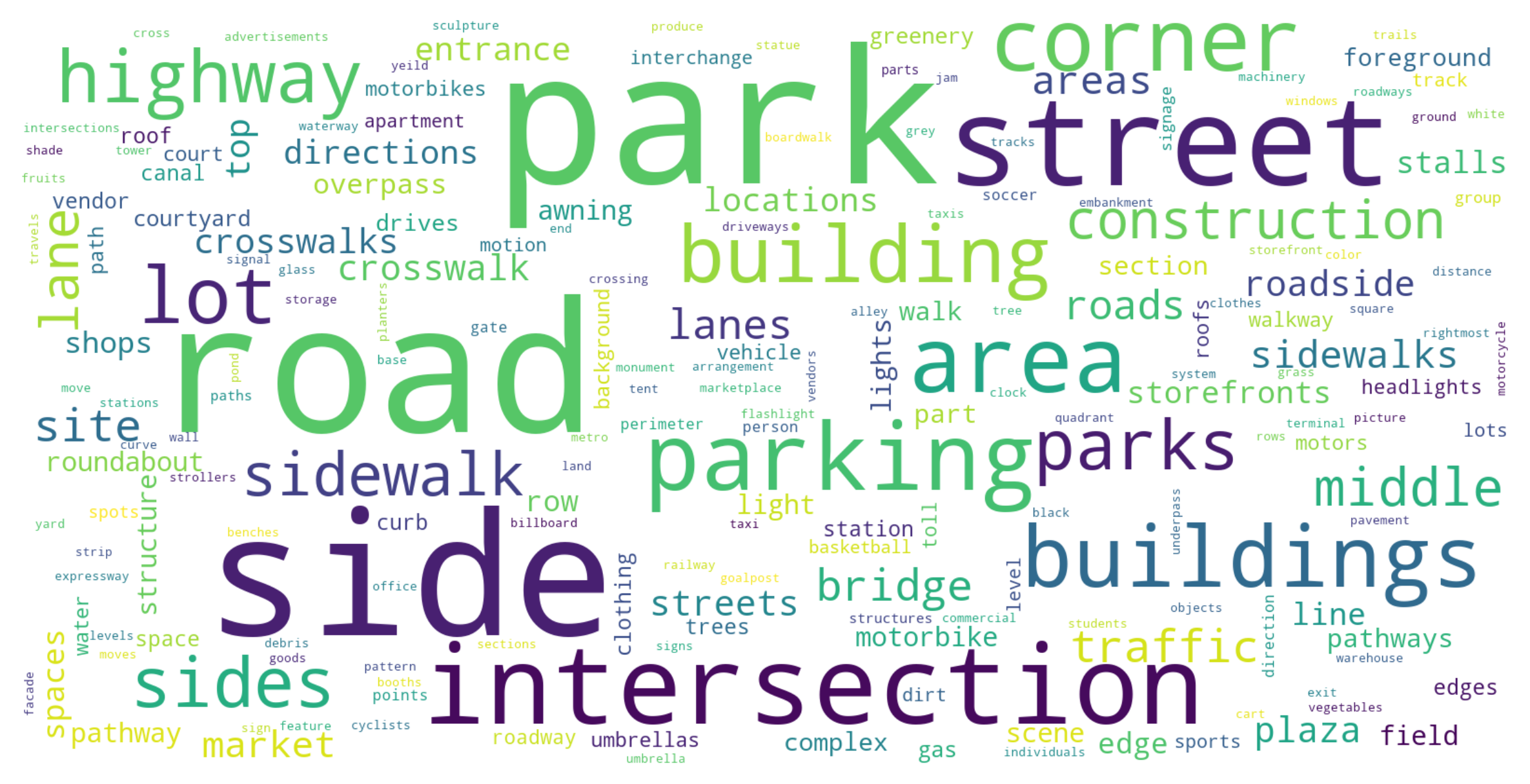}
    \end{minipage}
    }
    \vspace{-5pt}
    \caption{Word frequency visualization in RefDrone dataset.}
    \label{fig:words_cloud}
\vspace{-15pt}
\end{figure}

\begin{figure*}[ht]
    \centering
    \includegraphics[width=0.85\linewidth]{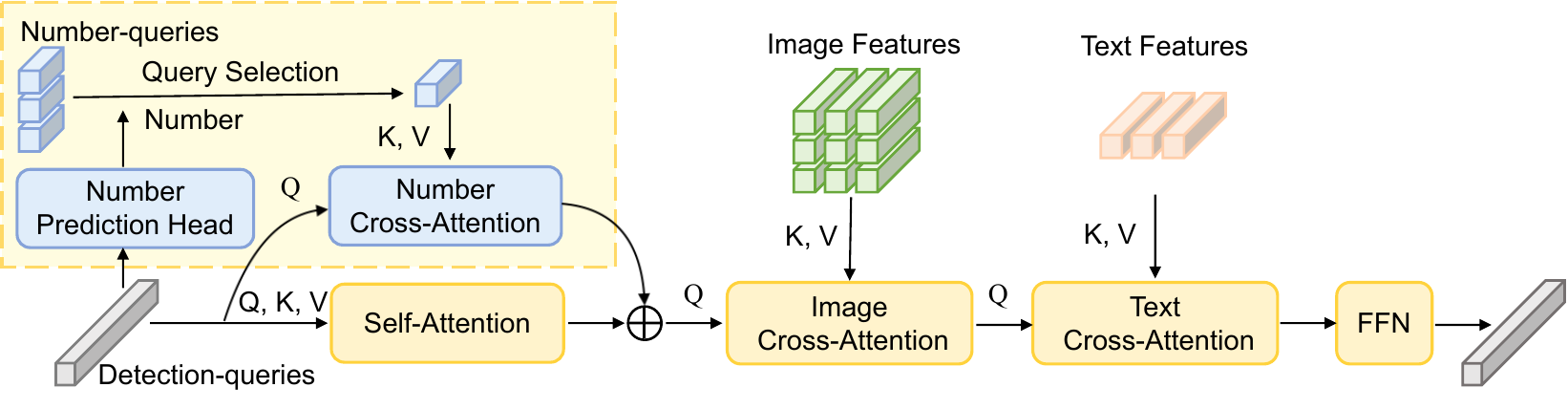}
\vspace{-10pt}
\caption{Architecture of a single decoder layer in Number GroundingDINO. Key modifications from GDINO~\cite{gdino} (highlighted in yellow box) include: (1) a number prediction head (FFN) to estimate target count, (2) number-queries selected through the predicted number, (3) a number cross-attention between selected number-queries and detection-queries.}
    \label{fig:ngdino}
\vspace{-10pt}
\end{figure*}

\subsection{Dataset analysis}
The RefDrone dataset comprises 17,900 referring expressions for 63,679 object instances across 8,536 images, spanning 10 categories. The dataset preserves the original train, validation, and test splits from VisDrone2019-DET~\cite{visdrone}. On average, each expression is 9.0 words long and refers to 3.8 objects. RefDrone is characterized by three primary challenges, illustrated in Figure~\ref{fig:intro-figure}: \par

\noindent\textbf{1) Multi-target and no-target samples.} In contrast to datasets like RefCOCO~\cite{refcoco} that focus on single-object references, RefDrone features a significant portion of complex queries, including 11,362 multi-target and 847 no-target expressions. The number of targets per expression ranges from 0 to 242. As shown in Figure~\ref{fig:data_statistics}, this distribution presents a greater challenge than that of gRefCOCO~\cite{grefcoco}, where expressions typically refer to only one or two objects. \par

\noindent\textbf{2) Multi-scale and small-scale target detection.} The dataset exhibits a wide distribution of object scales (Figure~\ref{fig:data_area}): small objects ($< 32^2$ pixels) account for 31\%, medium objects ($32^2$-$96^2$ pixels) for 55\%, and large objects ($> 96^2$ pixels) for 14\%. The high variance in object scales, particularly the prevalence of small objects, underscores multi-scale and small-scale target detection challenges. \par

\noindent\textbf{3) Complex environment with rich contextual reasoning.} Images are captured in complex environments with diverse viewpoints, lighting, and dense backgrounds. Consequently, the referring expressions extend beyond simple attributes (\eg, color) and spatial relationships (\eg, `left of') to describe complex object-object interactions (\eg, `the white trucks carrying livestock') and object-environment interactions (\eg, `the white cars line up at the intersection'). This complexity is visualized in the word clouds in Figure~\ref{fig:words_cloud}. \par

\subsection{Comparison to existing datasets}
Table~\ref{tab:dataset_ann} situates RefDrone among existing REC datasets. Key distinguishing features of RefDrone are its high average number of targets per expression and the use of an LVLM annotation pipeline. This pipeline generates expressions with richer contextual details compared to those from template-based methods or human-only annotation~\cite{zhang2023multi3drefer}. Although RIS-CQ~\cite{riscq} also uses an LLM for generation, its process is decoupled from visual content, leading to expressions that can be linguistically complex but visually ambiguous. While RSVG~\cite{zhan2023rsvg} targets small objects, the descriptive quality of its expressions is limited. In contrast, RefDrone comprehensively integrates these challenges, establishing it as a challenging benchmark in REC tasks.

\subsection{Evaluation metrics} To properly assess performance on multi-target expressions, we introduce instance-level metrics alongside traditional image-level ones. Standard image-level metrics, which treat the entire set of predictions for an expression as a single entity, are insufficient for our task. They cannot granularly penalize a model for missing individual objects within a large group. Our proposed instance-level metrics address this limitation by providing a fine-grained assessment of a model's ability to localize each individual target. \par

\smallskip
\textbf{Instance-level metrics Acc$_{inst.}$ and F1$_{inst.}$:} These metrics evaluate performance at the individual object level. We match each predicted bounding box to a ground-truth (GT) box. A prediction with an IoU $\geq$ 0.5 with a GT box is a true positive (TP). Unmatched predictions are false positives (FP), and unmatched GT boxes are false negatives (FN). For ``no-target'' samples, the absence of any prediction is a true negative (TN), while any prediction is an FP. Accuracy and F1-score are then computed as: 
\begin{equation}
\text{Acc}_{inst.} = \frac{\text{TP} + \text{TN}}{\text{TP} + \text{TN} + \text{FP} + \text{FN}},   \label{eq:important} 
\end{equation}
\vspace{-5pt}
\begin{equation}
\text{F1}_{inst.} = \frac{2 \cdot \text{TP}}{2 \cdot \text{TP} + \text{FP} + \text{FN}} .   \label{eq:also-important}
\end{equation}

\textbf{Image-level metrics Acc$_{img.}$ and F1$_{img.}$:} These metrics assess performance on the entire expression, enforcing a stricter success criterion. For a given expression, a prediction is a true positive (TP) only if the set of predicted boxes perfectly matches the set of ground-truth boxes. Any mismatch (e.g., missing, extra, or inaccurate boxes) results in a false positive (FP). TN and FP for ``no-target'' samples are defined as above. The metrics are calculated using the same formulas but on an image-wide basis.

\section{NGDINO}
We introduce Number GroundingDINO (NGDINO), a novel architecture designed to address the challenges of multi-target and no-target referring expression comprehension (REC). Our core insight is that explicit numerical reasoning about target counts is crucial for accurately grounding such expressions. NGDINO builds upon the strong foundation of GDINO~\cite{gdino}, inheriting its dual-encoder, single-decoder structure. Our contributions are concentrated within the decoder, as highlighted in Figure~\ref{fig:ngdino}, where we introduce three key components: (1) a number prediction head, (2) learnable number-queries with number-guided query selection, and (3) a number cross-attention module.

\smallskip
\noindent\textbf{Number prediction head.} To explicitly reason about the number of targets, we append a lightweight number prediction head to the decoder. This head takes the detection queries $Q_{det}$ as input, processes them through a Feed-Forward Network (FFN), and applies mean pooling to produce a probability distribution over target counts:
\begin{equation} 
    N_{prob} = \text{softmax} \left( \text{MeanPool} \left( \text{FFN}(Q_{det}) \right) \right), 
\end{equation} 
\vspace{-10pt}
\begin{equation} 
    N_{pred} = \text{argmax}(N_{prob}), 
\end{equation} 
where $Q_{det} \in \mathbb{R}^{B\times L_d \times D} $ are the detection queries for a  batch size $B$, query length $L_d$, and feature dimension $D$. To manage the long-tailed, Zipfian-like distribution of target counts~\cite{axtell2001zipf}, we discretize the output space into five bins: $\mathcal{C}=\{0,1,2,3,4+ \}$, where $4+$ represents four or more targets. This approach transforms the open-ended regression problem into a stable multi-class classification task.

\smallskip
\noindent\textbf{Number-guided query selection.} 
 To inject numerical priors into the decoding process, we introduce learnable number-queries, $Q_{num} \in \mathbb{R} ^{B\times L_n \times D}$, where $L_n$ is the length of number-queries. These queries are trained to encode distinct numerical patterns. The predicted number, $N_{pred}$, then acts as an index to select a dedicated slice of these queries: 
\begin{equation} 
Q_{num}^{sel} = Q_{num} \big[ :,\ L_s \cdot N_{pred}\ : \ L_s \cdot (N_{pred}+1),\ : \ \big], 
\end{equation}
where $L_s$ is the length of number-queries per category. This mechanism creates a mapping between the predicted cardinality and queries encoding patterns.

\smallskip
\noindent\textbf{Number cross-attention module.}
The selected number-queries, $Q_{num}^{sel}$, are fused with the detection queries, $Q_{det}$, via a number cross-attention module. This module operates in parallel with the self-attention layer within the decoder. The detection queries serve as the query (Q), while the selected number-queries serve as both key (K) and value (V). The output of this module is added to the output of the self-attention layer, enriching the detection queries with numerical context before they are passed to subsequent layers.

\smallskip
\noindent\textbf{Training objective.} The model is trained end-to-end by augmenting the GDINO losses (for bounding box regression and label classification) with a Cross-Entropy loss for the number prediction task. Hyperparameters, including the selected number-query length ($L_s =10$) and total number-query length ($L_n= 50$, representing 5 categories $\times$ 10 queries), were determined via ablation studies in the Appendix.

\section{Experiments}
We establish a comprehensive benchmark comprising 26 representative methods capable of performing REC tasks: 3 specialized REC methods, 7 task-specific LVLMs for REC, 12 general LVLMs with varying parameter scales, and 4 closed-source API models. To evaluate our proposed NGDINO, we conduct experiments on our RefDrone dataset and two public benchmarks: gRefCOCO~\cite{grefcoco} and RSVG~\cite{zhan2023rsvg}.

\begin{table*}[t]
\caption{Experimental results of zero-shot baselines on RefDrone benchmark. The best results in each group are denoted with \textbf{bold}.}
\vspace{-8pt}
  \centering
  \label{tab:zeroshot}
\begin{adjustbox}{width=\linewidth}
\begin{threeparttable}
  \begin{tabular}{@{}c l c  c  c c c c }
    \toprule
 \textbf{Categories} &    \textbf{Methods}  & \textbf{Params} & \textbf{Time} & \textbf{F1$_{inst.}$} &\textbf{ Acc$_{inst.}$} & \textbf{F1$_{img.}$} & \textbf{Acc$_{img.}$} \\
    \midrule 
  \multirow{4}{*}{\makecell{\textbf{Specialized} \\ \textbf{REC Methods}}}  &  $\text{MDETR}_\text{ResNet101}$~\cite{mdetr}  & 0.19B  & 2021/04/26  & \textbf{8.42} & \textbf{4.41} & 2.99 & 1.63 \\
 &  $\text{GLIP}_\text{Swin-Tiny}$~\cite{glip}    & 0.15B & 2021/12/06  & 5.46 & 3.84 & \textbf{9.20 }& \textbf{8.54} \\

 &  $\text{GroundingDINO}_\text{Swin-Tiny}$~\cite{gdino}  & 0.17B & 2023/05/09  & 1.18 & 1.84 & 3.94 & 6.35 \\

 &  $\text{GroundingDINO}_\text{Swin-Base}$~\cite{gdino}   & 0.23B & 2023/05/09   & 1.97 & 2.23 & 6.43 & 7.58 \\

\midrule

\multirow{7}{*}{\makecell{\textbf{Specialized } \\ \textbf{ LVLMs for REC }}}  & Shikra$^{\ddagger}$~\cite{chen2023shikra} & 7B & 2023/07/03 & 0.80 & 0.52 & 2.26 & 1.60 \\
& $\text{ONE-PEACE}_\text{Grounding}$$^{\ddagger}$
~\cite{one_peace} & 4B & 2023/07/20 & 1.02 & 0.51 & 2.64 & 1.34 \\
& Kosmos-2~\cite{peng2023kosmos} & 1.6B & 2023/10/30 & 8.06 & 4.20 & 8.64 & 4.52 \\
& $\text{CogVLM}_\text{Grounding}$~\cite{wang2023cogvlm} & 7B & 2023/11/20 & 15.38  & 8.33  & 30.73  & 18.15 \\
& Griffon~\cite{zhan2025griffon} & 13B & 2023/12/06 &  9.16  &  4.81  & 16.77   & 9.18   \\
& Ferret~\cite{you2023ferret}& 7B & 2023/12/14 & 3.18 & 1.62 & 8.48 & 4.43 \\

& Rex-Omni~\cite{jiang2025detect}  & 3B & 2025/10/15 & \textbf{54.06} & \textbf{37.10}  & \textbf{43.90}  & \textbf{28.52} \\

\midrule
\multirow{3}{*}{\makecell{\textbf{General LVLMs} \\ \textbf{ 0 \textasciitilde{} 7B}}} 
& $\text{DeepSeek-VL2}_\text{Tiny}$~\cite{wu2024deepseekvl2mixtureofexpertsvisionlanguagemodels} & 3B & 2024/12/13 & 2.35 & 1.84 & 4.51 & 4.98 \\

& Qwen2.5-VL~\cite{bai2025qwen2} & 3B &  2025/02/20 & 40.00 & 25.06 & 38.22  & 23.89  \\
& Qwen3-VL~\cite{qwen3technicalreport}  & 4B & 2025/10/15 & \textbf{48.68} & \textbf{32.31}  & \textbf{47.78}  & \textbf{32.28} \\

\midrule
\multirow{9}{*}{\makecell{\textbf{General LVLMs} \\ \textbf{  7B \textasciitilde{} 10B}}}   & Qwen-VL~\cite{bai2023qwen} & 7B & 2023/08/12  & 10.91 & 5.77 & 18.32 & 10.08 \\
& MiniGPT-v2~\cite{chen2023minigpt} & 7B & 2023/10/13   & 2.69 & 1.36 & 6.38 & 3.29 \\
& InternVL2.5$^{\ddagger}$~\cite{internvl2_5}  & 8B & 2024/12/05 & 0.58 & 1.14  & 1.79  & 4.2 \\
& Qwen2.5-VL~\cite{bai2025qwen2} & 7B &  2025/02/20 & 42.68	 & 27.20 & 	41.46	 & 26.52\\

& GLM 4.1V~\cite{vteam2025glm45vglm41vthinkingversatilemultimodal}  & 9B & 2025/07/01 & 24.39 & 14.05  & 30.57  & 18.74 \\
& Ovis2.5~\cite{lu2025ovis2} & 9B & 2025/08/19 & 4.24 & 2.20  & 14.56 & 8.39  \\
& $\text{MiMo-VL}_\text{RL}$~\cite{coreteam2025mimovltechnicalreport}  & 7B & 2025/08/21 & 19.75 & 11.22  & 30.24  & 18.79 \\

& InternVL3.5$^{\ddagger}$~\cite{wang2025internvl3_5}  & 8B & 2025/08/26 & 8.17 & 4.48  & 19.73  & 11.8 \\
& Qwen3-VL~\cite{qwen3technicalreport}  & 8B & 2025/10/15 & \textbf{51.66} & \textbf{35.13}  & \textbf{46.36 } & \textbf{31.85} \\

\midrule

\multirow{3}{*}{\makecell{\textbf{General LVLMs} \\ \textbf{  10B \textasciitilde{} 30B}}}   & SPHINX-v2$^{\ddagger}$~\cite{lin2023sphinx} & 13B & 2023/11/17  & 1.59 & 0.80  & 4.61 & 2.36 \\
& $\text{DeepSeek-VL2}_\text{Small}$~\cite{wu2024deepseekvl2mixtureofexpertsvisionlanguagemodels} & 16B & 2024/12/13 & 34.95 & 21.34	& 36.15 &  22.81   \\

& DeepSeek-VL2~\cite{wu2024deepseekvl2mixtureofexpertsvisionlanguagemodels} & 27B & 2024/12/13 & 25.31 & 14.88  & 29.46  & 19.08 \\

& Qwen3-VL~\cite{qwen3technicalreport} & 30B &  2025/10/15 & \textbf{56.88} & \textbf{39.98}  & \textbf{51.09}  & \textbf{35.58} \\

\midrule
\multirow{2}{*}{\makecell{\textbf{General LVLMs} \\ \textbf{ > 30B }}}  & GLM 4.5V~\cite{vteam2025glm45vglm41vthinkingversatilemultimodal}  & 106B & 2025/08/11 & 35.95 & 21.97  & 40.21  & 25.41 \\

& \cellcolor{gold!30}Qwen3-VL$^{\dagger}$~\cite{qwen3technicalreport} & \cellcolor{gold!30}235B & \cellcolor{gold!30}2025/10/04 & \cellcolor{gold!30}\textbf{58.79} & \cellcolor{gold!30}\textbf{41.93} & \cellcolor{gold!30}\textbf{52.16} & \cellcolor{gold!30}\textbf{36.89} \\

\midrule
\multirow{4}{*}{\makecell{\textbf{Closed-Source} \\ \textbf{(APIs) Models} }}  & DINO-XSeek~\cite{ren2024dino} & --   & 2025/03/11  & 54.47 & 37.46  & 46.11  & 30.14 \\

& Gemini 2.5 Pro~\cite{comanici2025gemini}  & -- & 2025/03/25 & 3.45 & 1.91  & 8.11  & 4.90 \\

& Seed1.5-VL~\cite{guo2025seed1}  & -- & 2025/05/12 & 43.52 & 27.84  & 37.27  & 23.03 \\
& Qwen3-VL-Plus~\cite{qwen3technicalreport}  & -- & 2025/09/22 & \textbf{58.11} & \textbf{41.32}  & \textbf{50.98}  & \textbf{36.18} \\
\bottomrule

\end{tabular}

{\footnotesize

\begin{tablenotes}
    \item[${\dagger}$] State-of-the-art method.  \quad ${\ddagger}$ Models only predict a single bounding box, limiting multi-target performance.
    \item[*] GPT and Claude are excluded due to output bounding-box format failures.
\end{tablenotes}
}
\end{threeparttable}
\vspace{-10pt}
\end{adjustbox}
\end{table*}

\subsection{Zero-shot results.}
As presented in Table~\ref{tab:zeroshot}, we evaluate all models in a zero-shot setting to assess their generalization capabilities to the drone scenes.

\vspace{3pt}
\noindent\textbf{Overall performance.} Among all models, the open-source Qwen3-VL (235B)~\cite{qwen3technicalreport} achieves state-of-the-art performance, with 58.79\% F1$_{inst.}$, 41.93\% Acc$_{inst.}$, 52.16\% F1$_{img.}$ , and 36.89\% Acc$_{img.}$. These results surpass not only other open-source models but also the specialized closed-source API, DINO-XSeek~\cite{ren2024dino} (54.47\% F1$_{inst.}$), which is explicitly designed for REC tasks.

\vspace{3pt}
\noindent\textbf{Analysis of specialized methods.} Specialized REC models (\eg, MDETR~\cite{mdetr}, GLIP~\cite{glip}, GroundingDINO~\cite{gdino}) show limited zero-shot transfer to RefDrone, with F1$_{inst.}$ generally below 10\%. This likely stems from pre-training on general-domain datasets, which restricts their cross-domain adaptability. Task-specific LVLMs for REC show varied performance.  Models limited to predicting a single bounding box, like Shikra~\cite{chen2023shikra}, ONE-PEACE~\cite{one_peace}, and SPHINX-v2~\cite{lin2023sphinx}, fail in multi-target scenarios, resulting in F1$_{inst.}$ scores below 2\%. In contrast, recent Rex-Omni~\cite{jiang2025detect} achieves competitive performance (54.06\% F1$_{inst.}$ / 43.90\% F1$_{img.}$), approaching the state-of-the-art.

\vspace{3pt}
\noindent\textbf{Analysis of general LVLMs.} A strong correlation is observed between model scale and zero-shot REC performance in general-purpose LVLMs. For instance, the Qwen3-VL series shows consistent improvement with scale: the 4B model achieves 48.68\% F1$_{inst.}$, which increases to 51.66\% for the 8B model and 56.88\% for the 30B model. This scaling trend suggests that larger model capacity directly enhances localization abilities without task-specific tuning. A notable exception is DeepSeek-VL2~\cite{wu2024deepseekvl2mixtureofexpertsvisionlanguagemodels}, where the small variant achieves better performance.

\vspace{3pt}
\noindent\textbf{Performance of closed-source APIs.} Commercial models do not consistently outperform open-source alternatives. DINO-XSeek~\cite{ren2024dino} achieves 54.47\% F1$_{inst.}$, lagging behind Qwen3-VL-235B. Other prominent APIs, including Gemini 2.5 Pro, yield low scores (F1$_{inst.}$ < 4\%). Furthermore, models from the GPT and Claude families failed to produce outputs in the required bounding-box format. This disparity indicates that many leading commercial models are not optimized for fine-grained localization tasks like REC.

\begin{table}[ht]
\caption{Results of fine-tuning baselines on RefDrone benchmark.}
  \centering
\vspace{-8pt}
  \begin{adjustbox}{width=\linewidth}
  \begin{tabular}{@{}l@{\hspace{4pt}}c  c c c @{}}
    \toprule
    Methods & F1$_{inst.}$ & Acc$_{inst.}$ & F1$_{img.}$ & Acc$_{img.}$ \\
    \midrule 

GLIP-T~\cite{glip} & 56.92 & 40.39  & 41.31 & 28.88  \\
GDINO-T~\cite{gdino}  & 67.64 & 51.55 & 54.54 & 39.63 \\
NGDINO-T (Ours) &  71.11 & 55.52  & 56.51  & 41.20 \\
GDINO-B~\cite{gdino}  & 69.75 & 53.95 & 56.95 & 41.81 \\
NGDINO-B (Ours)&  \textbf{72.51} & \textbf{57.22} & \textbf{57.84}  & \textbf{42.54} \\

  \bottomrule

\end{tabular}
\label{tab:finetune}

\end{adjustbox}
\end{table}

\subsection{Fine-tuning results.}

\noindent\textbf{Performance on the RefDrone Dataset.} Table~\ref{tab:finetune} presents the fine-tuning performance across specialized REC methods for 50 epochs on RefDrone dataset. Our proposed NGDINO exhibits consistent gains over the GDINO baseline. Specifically, the Swin-Tiny models NGDINO-T surpasses GDINO-T by 3.47\%, 3.97\%, 1.97\%, and 1.57\% in F1$_{inst.}$, Acc$_{inst.}$, F1$_{img.}$, and Acc$_{img.}$. Our NGDINO-B achieves the highest performance: 72.51\% F1$_{inst.}$, 57.22\% Acc$_{inst.}$, 57.84\% F1$_{img.}$ , and 42.54\% Acc$_{img.}$, outperforming the state-of-the-art LVLM Qwen3-VL-235B by a significant margin. Beyond the performance advantage, NGDINO requires orders of magnitude fewer parameters than LVLMs, making it suitable for deployment on resource-constrained edge platforms such as drones.

\begin{table}[t]
    \centering
  \caption{Experimental results on gRefCOCO~\cite{grefcoco} dataset. Asterisk (*) denotes results reported in the gRefCOCO paper.  }\label{tab:results_grec}
  \vspace{-8pt}
  \begin{adjustbox}{width=\linewidth}
     \begin{tabular}{@{}l @{\hspace{2pt}}c @{\hspace{5pt}}c @{\hspace{5pt}}|c @{\hspace{5pt}}c@{}}
     \toprule
      \multirow{2}{*}{Methods} & \multicolumn{2}{c|}{testA} & \multicolumn{2}{c}{testB} \\
       &    Pr@0.5~$\uparrow$ &  N-acc.~$\uparrow$   &Pr@0.5~$\uparrow$    & N-acc.~$\uparrow$  \\
    \midrule

        MDETR*~\cite{mdetr} & \textbf{50.0} & 34.5 & 36.5 & 31.0   \\
        UNINEXT*~\cite{yan2023universal}  & 46.4 & 49.3 & 42.9 & 48.2 \\
        GDINO-T~\cite{gdino}  &  45.69 & 79.02  & 44.83  & 76.69 \\
        NGDINO-T & 46.05  & \textbf{83.17}  & \textbf{45.57} & \textbf{78.08} \\
        \bottomrule
\end{tabular}
\end{adjustbox}
\end{table}

\vspace{3pt}
\noindent\textbf{Performance on public benchmarks.} To further validate the effectiveness of our improvements over GDINO in handling multi-target scenarios, we evaluate NGDINO on two benchmarks that contain multi-target samples: gRefCOCO~\cite{grefcoco} and RSVG~\cite{zhan2023rsvg}. Models are fine-tuned for 5 epochs on both benchmarks. Table~\ref{tab:results_grec} presents results on gRefCOCO~\cite{grefcoco}, which includes both multi-target and no-target samples. We adopt two metrics: Pr@0.5, measuring the percentage of predictions achieving $F1$ = 1 at IoU $\geq$ 0.5, and N-acc, denoting accuracy on no-target samples. NGDINO-T demonstrates improvements over GDINO-T in no-target detection, with N-acc gains of 4.15\% and 1.39\% on test A and test B, respectively.  The improvements in Pr@0.5 are more modest (0.36\% and 0.74\% on test A and B), which can be attributed to the relatively simple multi-target structure in gRefCOCO, where expressions predominantly reference only one or two objects.  While MDETR~\cite{mdetr} achieves higher Pr@0.5 on test A, this comes at the cost of lower N-acc due to its tendency to produce excessive outputs, leading to false positives on no-target samples. On the more complex aerial-view RSVG~\cite{zhan2023rsvg} benchmark (Table~\ref{tab:results_rsvg}), NGDINO-T consistently outperforms prior state-of-the-art methods across all metrics. These results confirm that our proposed modifications effectively enhance performance on diverse multi-target REC tasks.

\begin{figure}[t]
    \centering
    \includegraphics[width=1\linewidth]{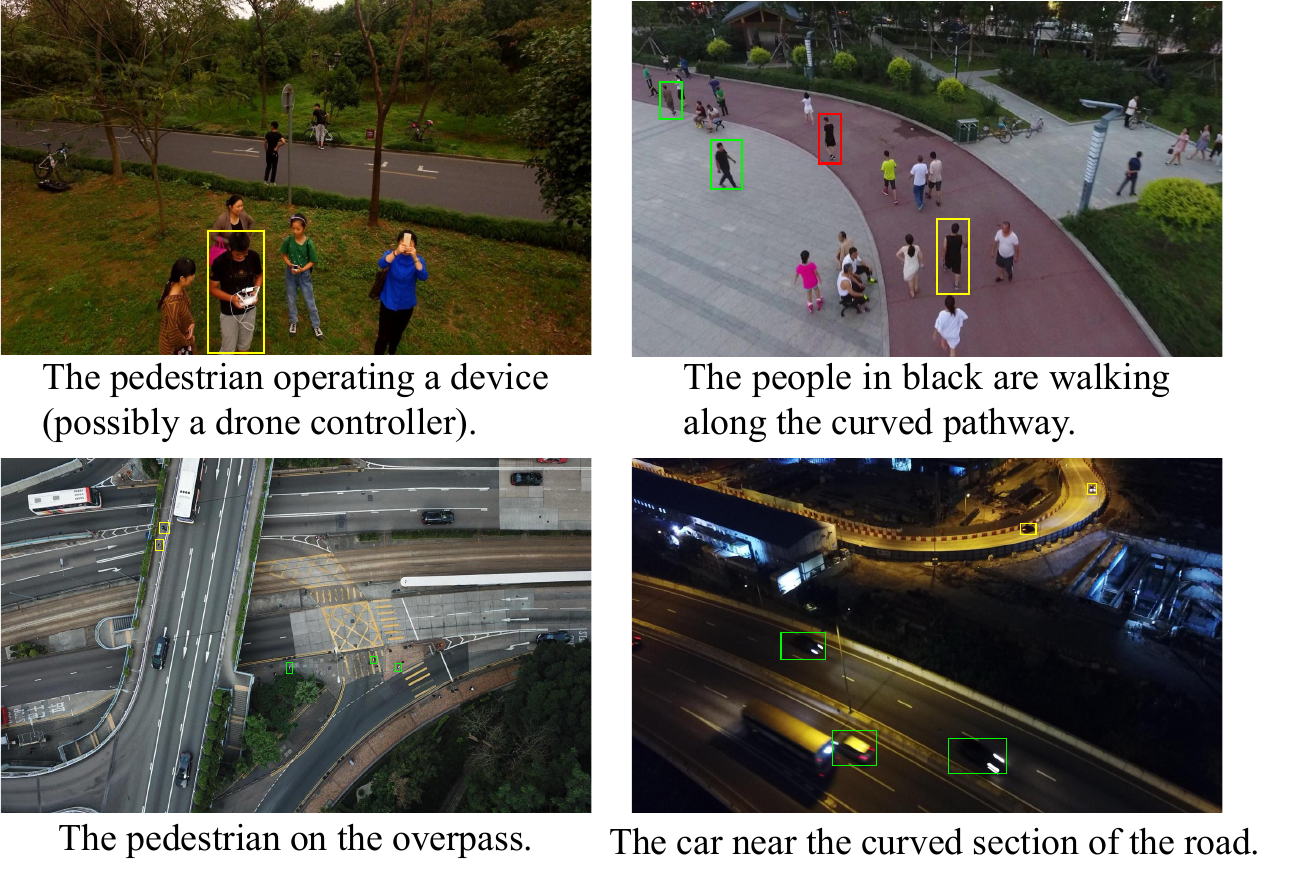}
    \vspace{-8pt}
\caption{Some failure cases of NGDINO on RefDrone dataset. Red, green, and yellow boxes indicate true positives, false positives, and false negatives, respectively.}
    \label{fig:fail}
\end{figure}

\subsection{Ablation studies}
\textbf{Analysis of NGDINO components.} Table~\ref{tab:ngdino_ablation} presents the analysis of each component in NGDINO. With only the number prediction head, the model achieves improvements, improving F1$_{inst.}$ from 67.64\% to 69.73\% (+2.09\%). This demonstrates that the auxiliary number prediction task enhances the model's grounding capability. The number cross-attention mechanism alone yields more modest gains, improving F1$_{inst.}$ to 68.88\% (+1.24\%). This improvement can be partially attributed to the additional parameters introduced in the decoder. The most significant performance boost occurs when combining both components, where F1$_{inst.}$ reaches 71.11\% (+3.47\%). This substantial accuracy improvement comes at a negligible cost to efficiency, with only a minor decrease in inference speed (from 13.5 to 12.3 FPS). \par

\smallskip
\noindent\textbf{Number prediction accuracy.} We evaluate the number prediction head's performance. The head achieves a mean absolute error (MAE) of 0.21, indicating that its numerical predictions are highly precise and closely align with the ground truth counts. This strong performance in number prediction is a key factor in the model's ability to handle multi-target expressions. The overall accuracy for predicting the correct number of instances is 75.3\%. \par

\begin{table}[t]
  \caption{Experimental results on RSVG~\cite{zhan2023rsvg} dataset. Asterisk (*) denotes results reported in the EarthGPT paper. Pr@0.5: percentage of predictions at IoU $\geq$ 0.5.. }\label{tab:results_rsvg}
  \vspace{-8pt}
\begin{adjustbox}{width=\linewidth}
     \begin{tabular}{@{}l@{\hspace{1pt}}c@{\hspace{4pt}}c@{\hspace{4pt}} c@{\hspace{4pt}} c@{\hspace{4pt}}c@{}}
     \toprule
     Methods  &  Pr@0.5 &  Pr@0.6   &Pr@0.7    & Pr@0.8  & Pr@0.9 \\
    \midrule

        TransVG*~\cite{deng2021transvg} & 72.41 & 67.38 & 60.05 & 49.10 & 27.84 \\
        MGVLF*~\cite{zhan2023rsvg} & 76.78 & 72.68 & 66.74 & 56.42 & 35.07 \\
        EarthGPT*~\cite{zhang2024earthgpt} & 76.65 & 71.93 & 66.52 & 56.53 & 37.63 \\

        GDINO-T~\cite{gdino}  & 76.99 & 75.81 & 73.36 & 66.77 & 49.96\\
        NGDINO-T  & \textbf{77.16} & \textbf{76.03} & \textbf{73.81} & \textbf{67.84} & \textbf{51.39} \\
        \bottomrule
\end{tabular}

\end{adjustbox}

\end{table}
\begin{table}[t]
\caption{Ablation study of NGDINO components on the RefDrone dataset with Swin-Tiny backbone. }
\vspace{-8pt}
  \centering
\begin{adjustbox}{width=\linewidth}
  \begin{tabular}{@{}c@{\hspace{1pt}} |@{\hspace{1pt}} c@{\hspace{1pt}}| @{\hspace{1pt}}c @{\hspace{1pt}}c @{\hspace{1pt}}c @{\hspace{1pt}}c @{\hspace{1pt}}c@{}}
    \toprule

   \multirow{2}{*}{\makecell{{Number} \\ {Prediction Head}}} & \multirow{2}{*}{\makecell{{Number} \\ {Cross-Attention}}}  & \multirow{2}{*}{F1$_{inst.}$} & \multirow{2}{*}{Acc$_{inst.}$}  &  \multirow{2}{*}{F1$_{img.}$}  & \multirow{2}{*}{Acc$_{img.}$} & \multirow{2}{*}{FPS}  \\
    & & & & & & \\
    \midrule 

         &         & 67.64 & 51.55 & 54.54 & 39.63  & \textbf{13.5} \\
     \checkmark &    &  69.73 & 53.90 & 56.29 &  41.08  & 12.8  \\
  &  \checkmark   &  68.88 &  52.89   & 54.98  & 39.74  & 12.9   \\
 \checkmark & \checkmark & \textbf{71.11} & \textbf{55.52} & \textbf{56.51} & \textbf{41.20}  & 12.3\\

  \bottomrule

\end{tabular}
\end{adjustbox}
\label{tab:ngdino_ablation}
\end{table}

\subsection{Limitations}
Despite its strong performance on multi-target and no-target scenarios, NGDINO has several limitations. As illustrated by the qualitative examples in Figure~\ref{fig:fail}, failure cases typically arise from challenging scenarios inherent to the RefDrone dataset. These failures can be categorized into three primary sources: (1) complexity in the referring expression that demands sophisticated contextual reasoning; (2) cluttered backgrounds that camouflage target objects; and (3) inherent difficulties in detecting very small-scale objects. Addressing these challenging, real-world conditions remains an important direction for future work.

\section{Conclusion}
In this work, we introduce RefDrone, a challenging benchmark for referring expression comprehension in drone scenes. The dataset is built using RDAnnotator, an efficient semi-automated annotation pipeline that combines LVLMs with human-in-the-loop verification to ensure high-quality annotations. Our extensive experiments reveal a substantial performance drop for existing state-of-the-art methods when evaluated on RefDrone, underscoring the benchmark's difficulty and exposing key limitations in current REC approaches. Furthermore, we develop NGDINO to address the multi-target and no-target challenges in RefDrone. In the future, we aim to further enhance NGDINO to address additional challenges presented by RefDrone. We also plan to extend RefDrone to larger-scale scenarios and introduce additional tasks such as referring expression segmentation and tracking. We believe RefDrone will serve as a valuable benchmark for advancing research in drone-based REC tasks.

{
\small
\bibliographystyle{plain}
\bibliography{main}

}

\clearpage

\appendix
\section{Appendix}

\noindent We provide the following appendices for further analysis:
\begin{compactitem}
    \item Details of the baseline methods. (Appendix~\ref{sup:baseline})
    \item Implementation details. (Appendix~\ref{sup:implementation})
    \item Details of our color categorization pipeline. \\ (Appendix~\ref{sup:color})
    \item Results with respect to different object scales. \\ (Appendix~\ref{sup:scale})
    \item Ablation study on query length. (Appendix~\ref{sup:ablation_length})
    \item Performance validation of the RDAnnotator framework. (Appendix~\ref{sup:ablation_rdannotator})
    \item Results on standard RefCOCO/+/g datasets. \\ (Appendix~\ref{sup:refcoco})
    \item Additional examples from the RefDrone dataset. \\ (Appendix~\ref{sup:images})
    \item Prompts and examples used in RDAnnotator.  \\ (Appendix~\ref{sup:prompt})
\end{compactitem}

\subsection{Details of baseline methods}
\label{sup:baseline}

The details for each baseline method:
\begin{compactitem} 
\item \textbf{MDETR}: ResNet-101 with BERT-Base, \newline
pretrained on Flickr30k, RefCOCO/+/g, VG.
\item \textbf{GLIP}: Swin-Tiny with BERT-Base, \newline
pretrained on Objects365.
\item \textbf{GDINO-T}: Swin-Tiny with BERT-Base, \newline
pretrained on Objects365, GoldG, GRIT, V3Det.
\item \textbf{GDINO-B}: Swin-Base with BERT-Base, \newline
pretrained on Objects365, GoldG, V3Det.
\end{compactitem}

\subsection{Implementation details}
\label{sup:implementation}
\noindent\textbf{NGDINO implementation.} We train NGDINO using a two-stage procedure to ensure stability. Stage 1: Pre-training. We initialize the model with weights from a pre-trained GDINO~\cite{gdino}, freezing all components except for the number prediction head for 5 epochs. This new head is then pre-trained on the RefDrone dataset. Stage 2: End-to-end Fine-tuning. After the head is pre-trained, we unfreeze the entire model and fine-tune it on our target dataset. This staged approach prevents the randomly initialized prediction head from destabilizing the well-trained detector backbone during the initial phases of training.

\smallskip
\noindent\textbf{Zero-shot evaluation protocol.} For all baseline models, we adhere to established evaluation practices to ensure fair comparisons. \textbf{Specialized REC Methods}: For GLIP~\cite{glip} and GDINO~\cite{gdino}, we use the official model checkpoints and implementations provided within the MMDetection~\cite{mmdetection} framework. \textbf{LVLMs:}  For a standardized and reproducible evaluation of LVLMs, we integrate our dataset into the VLMEvalKit framework~\citep{duan2024vlmevalkit}. Within this framework, we benchmark each model using its official, recommended prompt structure to ensure optimal performance.
\par

\smallskip
\noindent\textbf{Fine-tuning evaluation details.} All fine-tuning experiments are conducted within the MMDetection~\cite{mmdetection} framework on 8 NVIDIA A100 GPUs. To ensure a fair comparison, we apply a consistent protocol across all models. We follow the original learning strategies and hyperparameter settings for each model with one critical modification: we disable random crop data augmentation. This is because random cropping can remove crucial spatial context or the target objects themselves in our position-sensitive referring expressions, introducing label noise and degrading performance.

\subsection{Details of color categorization.}
\label{sup:color}
Color is a foundational attribute in the RefDrone dataset, present in 69\% of all referring expressions. However, accurately identifying color is non-trivial due to challenges like illumination variance, occlusions, and semantic ambiguity (e.g., distinguishing "red" from "pink" or "orange"). To address this, we designed a hybrid color extraction pipeline that combines the efficiency of a specialized classifier with the reasoning capabilities of an LVLM. The pipeline consists of two stages: 

 \noindent \textbf{(1) Classifier-based Proposal:} A WideResNet-101 classifier~\cite{wideresnet} generates an initial color prediction. To create a high-quality training set for this classifier, we first generate labels programmatically using the HSV color space and then perform manual validation to correct noise and refine ambiguous cases. 
 
  \noindent  \textbf{(2) LVLM Verification:} An LVLM verifier then assesses the classifier's output. Using structured prompts, it reasons about the visual evidence to confirm the prediction or correct it, effectively resolving ambiguities caused by lighting or partial visibility. 
 
\begin{figure}[h]
\centering
\includegraphics[width = 0.85\linewidth]{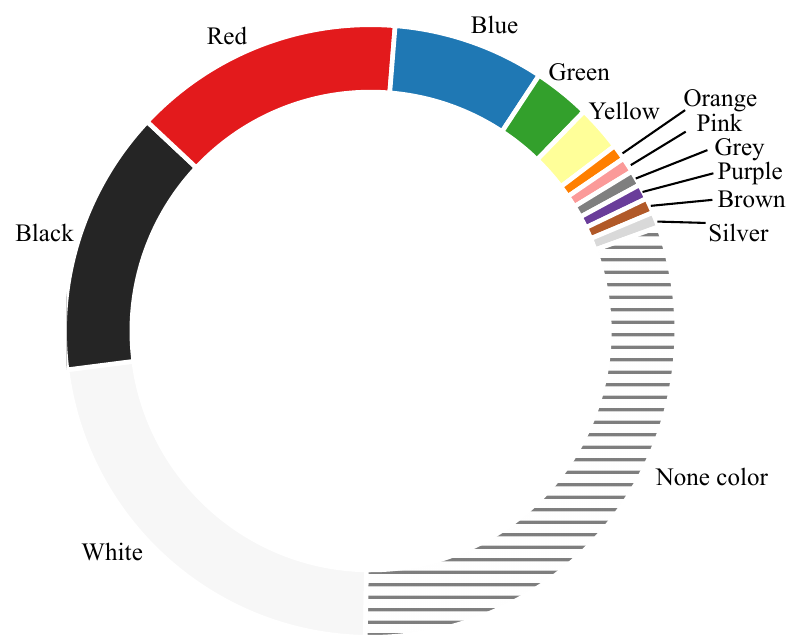}
\caption{Distribution of color terms in RefDrone expressions.} 
\label{fig:color}
\end{figure}

 The reliability of this hybrid approach enabled us to expand our vocabulary from an initial set of six primary colors (e.g., red, blue) to a more nuanced palette of twelve, including orange, pink, grey, and purple. The final distribution of these color terms is visualized in Figure~\ref{fig:color}.

\subsection{The results of different object scales.}
\label{sup:scale}
To provide a granular analysis of model robustness to scale variation, a critical challenge in  the RefDrone benchmark, we evaluated a representative set of high-performing methods (ACC$_{inst.}$ > 10\%) on objects categorized as small, medium, and large. The results, presented in Table~\ref{tab:sml}, reveal a stark performance gap between LVLMs and Specialized REC methods, particularly on small objects. Notably, even LVLMs like Qwen3-VL-235B struggle,  achieving only 26.54\%  Acc$_{s}$. In contrast, among the fine-tuned specialized REC methods, our NGDINO-B achieves 44.08\% Acc$_{s}$, 62.59\% Acc$_{m}$, and 68.20\% Acc$_{l}$.

\begin{table}[h]
\caption{Performance comparison on small (Acc$_{s}$), medium (Acc$_{m}$), and large (Acc$_{l}$) objects. The models in the upper section are LVLMs evaluated in a zero-shot setting. The models in the lower section are specialized REC methods fine-tuned on the RefDrone training set.}
\vspace{-5pt}
\centering
\begin{adjustbox}{width=\linewidth}

\begin{tabular}{@{}l@{\hspace{1pt}}c ccc c @{}}
    \toprule
    Methods & Params & Acc$_{s}$ & Acc$_{m}$ & Acc$_{l}$ & Acc$_{inst.}$ \\
    \midrule

    Rex-Omni~\cite{jiang2025detect}  & 3B  & 25.55 & 43.03 & 50.31 & 37.10\\
    
    Qwen2.5-VL~\cite{bai2025qwen2} & 3B &  13.52 & 28.59 & 32.44 & 25.06 \\
    Qwen2.5-VL~\cite{bai2025qwen2} & 7B &  12.19 & 33.36 & 35.80  & 27.20 \\

    Qwen3-VL~\cite{qwen3technicalreport} & 4B & 18.53 & 42.66  &  54.23 & 32.31 \\
    Qwen3-VL~\cite{qwen3technicalreport} & 8B & 20.05 & 43.64 &  50.32  & 35.13 \\
    Qwen3-VL~\cite{qwen3technicalreport} & 30B & 24.21 & 50.48  & 54.68  & 39.98 \\
    Qwen3-VL~\cite{qwen3technicalreport} & 235B & \textbf{26.54} &\textbf{ 51.60} & 55.79  & \textbf{41.93} \\

    $\text{MiMo-VL}_\text{RL}$~\cite{coreteam2025mimovltechnicalreport}  & 7B &  2.41 & 13.69  & 15.65  & 11.22\\
    
    GLM 4.1V~\cite{vteam2025glm45vglm41vthinkingversatilemultimodal}  & 9B & 1.25 & 16.94 & 29.84  & 14.05\\
    GLM 4.5V~\cite{vteam2025glm45vglm41vthinkingversatilemultimodal}  & 106B & 7.70 & 27.83 & 29.17  & 21.97 \\

    $\text{DeepSeek-VL2}_\text{Small}$ & 16B & 5.83 & 27.64  & 30.36 & 21.34 \\
    DeepSeek-VL2~\cite{wu2024deepseekvl2mixtureofexpertsvisionlanguagemodels} & 27B & 2.66 & 18.58  & 18.57 & 14.88 \\

    DINO-XSeek~\cite{ren2024dino} & --   & 24.54 & 46.52  & 62.51 & 37.46 \\
    Seed1.5-VL~\cite{guo2025seed1}  & -- & 11.11 & 38.39  & 50.83 & 27.84 \\
    Qwen3-VL-Plus~\cite{qwen3technicalreport}  & -- & 25.34 & 49.96 & \textbf{56.18}  & 41.32 \\

    \midrule

    GLIP-T~\cite{glip} & 0.15B  &  21.08 & 48.82 & 54.35 & 40.39 \\
    GDINO-T~\cite{gdino}  &  0.17B & 38.56 & 56.69 & 66.13 & 51.55 \\
    GDINO-B~\cite{gdino}  & 0.23B  & 40.01 & 59.96 & 67.94 & 53.95 \\
    NGDINO-T (Ours)  & 0.18B & 42.48 & 60.42 & 67.90  & 55.52\\
    NGDINO-B (Ours)  & 0.24B  & \textbf{44.08} & \textbf{62.59} & \textbf{68.20}  & \textbf{57.22} \\

    \bottomrule
\end{tabular}
\label{tab:sml}
\end{adjustbox}
\vspace{-5pt}
\end{table}

\subsection{Ablation study on query length.} 
\label{sup:ablation_length}
Table~\ref{tab:length_ablation} analyzes the impact of varying the query length. A minimal query length of 1 lacks the capacity to capture complex numerical information. Conversely, extending the query length to 100 increases parameter count and computational overhead, potentially leading to optimization challenges. Through these experiments, we determine that a query length of 10 provides an optimal trade-off.\par
\begin{table}[h]
    \centering
    \caption{Ablation study on the impact of varying selected number query length. Params indicates additional parameters introduced.}
    \vspace{-5pt}
    \label{tab:length_ablation}
\begin{adjustbox}{width=\linewidth}
  \begin{tabular}{l  @{\hspace{10pt}}| c  @{\hspace{10pt}} c @{\hspace{10pt}} c  @{\hspace{10pt}}c  @{\hspace{10pt}}c}
    \toprule
    Length &  F1$_{inst.}$ & Acc$_{inst.}$ & F1$_{img.}$ & Acc$_{img.}$ & Params\\
    \midrule 
1  & 70.20 & 54.44 & 55.82 & 40.56 & 1.58M\\
10  & \textbf{71.11} & \textbf{55.52} & \textbf{56.51} & \textbf{41.20} & 1.65M\\
100 &  70.44 & 54.72  & 55.95  & 40.73 &  2.34M\\
  \bottomrule
\end{tabular}
\end{adjustbox}

\end{table}

\subsection{Performance of the RDAnnotator framework} 
\label{sup:ablation_rdannotator}

To validate the effectiveness of our proposed annotation framework RDAnnotator, we evaluate RDAnnotator when repurposed as a complete, two-stage method for REC. To ensure a fair comparison, all methods operate on an identical set of initial object proposals generated by a first-stage Faster-RCNN detector~\cite{fasterrcnn} (18.0 mAP). The core of the evaluation lies in the second stage, where each method uses the referring expression to rank these proposals and identify the target. We benchmark RDAnnotator against two strong alternative second-stage approaches: (1) \textbf{GPT-4o}, which represents a powerful, single-step LVLM reasoning approach, and (2) \textbf{ReCLIP}~\cite{reclip}, a representative CLIP-based ranker that relies on embedding similarity. As shown in Table~\ref{tab:agent_ablation}, RDAnnotator substantially outperforms both, validating the efficacy of its structured, multi-step reasoning process. Results underscore RDAnnotator's suitability for generating high-fidelity REC annotations.

\begin{table}[h]

    \centering
    \caption{Experimental results of two-stage instance ranking methods on the RefDrone benchmark. }
    \vspace{-8pt}
\label{tab:agent_ablation}
\begin{adjustbox}{width=\linewidth}
  \begin{tabular}{ l  @{\hspace{12pt}} c  @{\hspace{12pt}} c @{\hspace{12pt}} c @{\hspace{12pt}} c  }
    \toprule
    Methods & F1$_{inst.}$ & Acc$_{inst.}$ & F1$_{img.}$ & Acc$_{img.}$ \\
    \midrule 
ReCLIP~\cite{reclip}  & 24.62 & 14.04 & 11.58 & 6.15 \\
GPT4-o  & 52.38 & 35.65 & 35.50 & 22.38 \\
RDAnnotator &  \textbf{58.14}  & \textbf{41.13} & \textbf{37.07}  & \textbf{23.54} \\

  \bottomrule

\end{tabular}
\end{adjustbox}
\end{table}

\subsection{Results on RefCOCO/+/g datasets.}
\label{sup:refcoco}
Since the RefCOCO, RefCOCO+, and RefCOCOg datasets contain only one instance per expression, the proposed NGDINO leverages the number branch primarily to address multi-instance and no-instance scenarios. As a result, the performance of NGDINO is relatively similar to that of GDINO on these datasets.

\begin{table}[h]
\caption{Results on RefCOCO/+/g datasets.}
\vspace{-5pt}
\centering
\begin{adjustbox}{width=\linewidth}
\begin{tabular}{@{}l@{\hspace{2pt}}|c@{\hspace{4pt}}c|c@{\hspace{4pt}}c|c@{\hspace{4pt}}c@{}}
\toprule
& \multicolumn{2}{c|}{RefCOCO} & \multicolumn{2}{c|}{RefCOCO+} & \multicolumn{2}{c}{RefCOCOg} \\
& TestA & TestB & TestA & TestB & Val & Test \\
\midrule
MDETR & 90.4 & 82.67 & 85.52 & 72.96 & 83.35 & 83.31\\
GDINO-T & 91.4 & \textbf{86.6} & 87.5 & 74.0 & \textbf{85.5} & 85.8\\
NGDINO-T & \textbf{91.5} & 86.5 & \textbf{87.8} & \textbf{74.7} & 85.3 & \textbf{85.8} \\
\bottomrule
\end{tabular}
\end{adjustbox}
\end{table}

\subsection{Dataset examples}
\label{sup:images}
\begin{figure*}[ht]
\centering
\includegraphics[width = 0.95\textwidth]{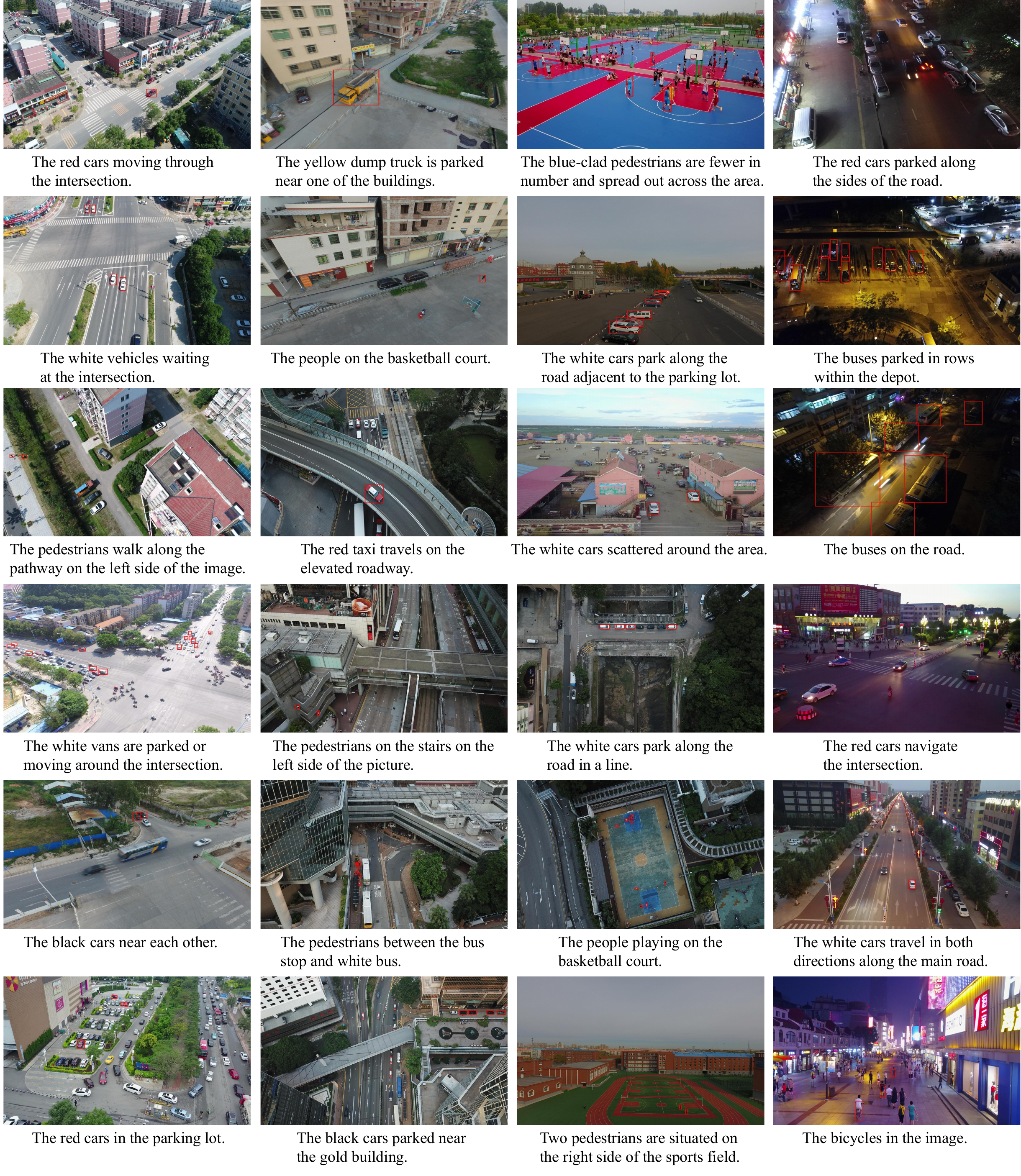}
\vspace{-8pt}
\caption{Dataset examples from RefDrone.} 
\vspace{-8pt}
\label{fig:example}
\end{figure*}

To provide a comprehensive understanding of our RefDrone dataset, we present representative examples in Figure~\ref{fig:example}. These samples demonstrate the three key challenges in our dataset, highlighting its real-world applicability.

\subsection{Prompts and examples for RDAnnotator}
\label{sup:prompt}
In this section, we provide the prompts and examples employed in RDAnnotator. Table~\ref{table9} presents the prompt construction process for expression generation (Step 3), which includes the system prompt and few-shot in-context learning examples. One in-context learning example is illustrated in Table~\ref{gpt-ass}. The system prompts used for each step are detailed in Table~\ref{table11}. Additionally, the system prompts for the feedback mechanism are presented in Table~\ref{table12}.

\begin{table*}[t]\centering
\caption{Illustration of RDAnnotator's prompt construction for expression generation (Step 3). Few-shot in-context-learning examples are from \VarSty{fewshot\_samples}. A representative example is shown in Table~\ref{gpt-ass}}
\begin{minipage}{0.99\textwidth}\vspace{0mm}    \centering
\begin{tcolorbox} 
    \centering
    \small
     \hspace{-6mm}
    \begin{tabular}{p{0.99\textwidth}}
\label{prompt construction}
\begin{minipage}{0.99\textwidth}\vspace{0mm}

\VarSty{messages} = [
\{\var{"role":"system", "content":} f\var{"""}As an AI visual assistant, your role involves analyzing a single image. You are supplied with three sentences that \textbf{caption} the image, along with additional data about specific attributes of objects within the image. This can include information about \textbf{categories, colors, and precise coordinates}. Such coordinates, represented as floating-point numbers that range from 0 to 1, are shared as center points, denoted as (x, y), identifying the center x and y. When coordinate x tends to 0, the object nears the left side of the image, shifting towards the right as coordinate x approaches 1. When coordinate y tends to 0, the object nears the top of the image, shifting towards the bottom as coordinate y approaches 1. \\
            
Your task is to classify the provided objects based on various characteristics, while also substantiating your classification. This classification should be thoroughly justified, with criteria including but not limited to relationships or relative locations of objects.\\

To refer to a specific object, use the provided coordinates directly. Base your classification justifications on direct observations from the image, avoiding any hypothesizing or assumptions.\var{"""}\}\\
        ]
        
    \For{ \VarSty{sample} in   \VarSty{fewshot\_samples}}{
         \var{\VarSty{messages}.append(\{"role":"user", "content":\VarSty{sample[`context']}\})} \; \\
         \var{\VarSty{messages}.append(\{"role":"assistant", "content":\VarSty{sample[`response']}\} ) } \;
         }  
    \var{\VarSty{messages}.append(\{"role":"user", "content":`\textbackslash  n'.join(\VarSty{query})\})}

\end{minipage}
    \end{tabular}
\end{tcolorbox}
    
\vspace{-2mm}

\end{minipage}

\label{table9}

\end{table*}

\begin{table*}[t]
\caption{One example to illustrate the input and output in RDAnnotator Step 3 (expression generation). The top block shows the input contexts such as image, object, caption and color attributed. The bottom block shows the generated annotations.}

\begin{tcolorbox}

\blue{\textbf{Context type 1: Captions}}\\
The image shows an aerial view of a busy urban street with multiple lanes of traffic, flanked by parking areas and buildings. A median strip with greenery runs down the center, dividing the opposite flows of vehicles. On either side of the road, there are commercial and residential structures. Parking lanes on both sides of the main road are filled with parked vehicles.\\
The image shows a bird's-eye view of an urban street scene with traffic congestion...$\langle\text{omitted}\rangle$ \\
\begin{minipage}{0.55\textwidth}
\blue{\textbf{Context type 2: Objects}} \\
pedestrian, white: [0.173, 0.849]  \\
car, black: [0.214, 0.975]  \\
car, white: [0.215, 0.904] \\
car, white: [0.313, 0.444] \\
car, red: [0.315, 0.256] \\
van, white: ...$\langle\text{omitted}\rangle$\\
\blue{\textbf{Context type 3: Image}} 
\end{minipage}
\begin{minipage}{0.4\textwidth}
    \centering
    \rotatebox{270} {\includegraphics[width=0.8\linewidth]{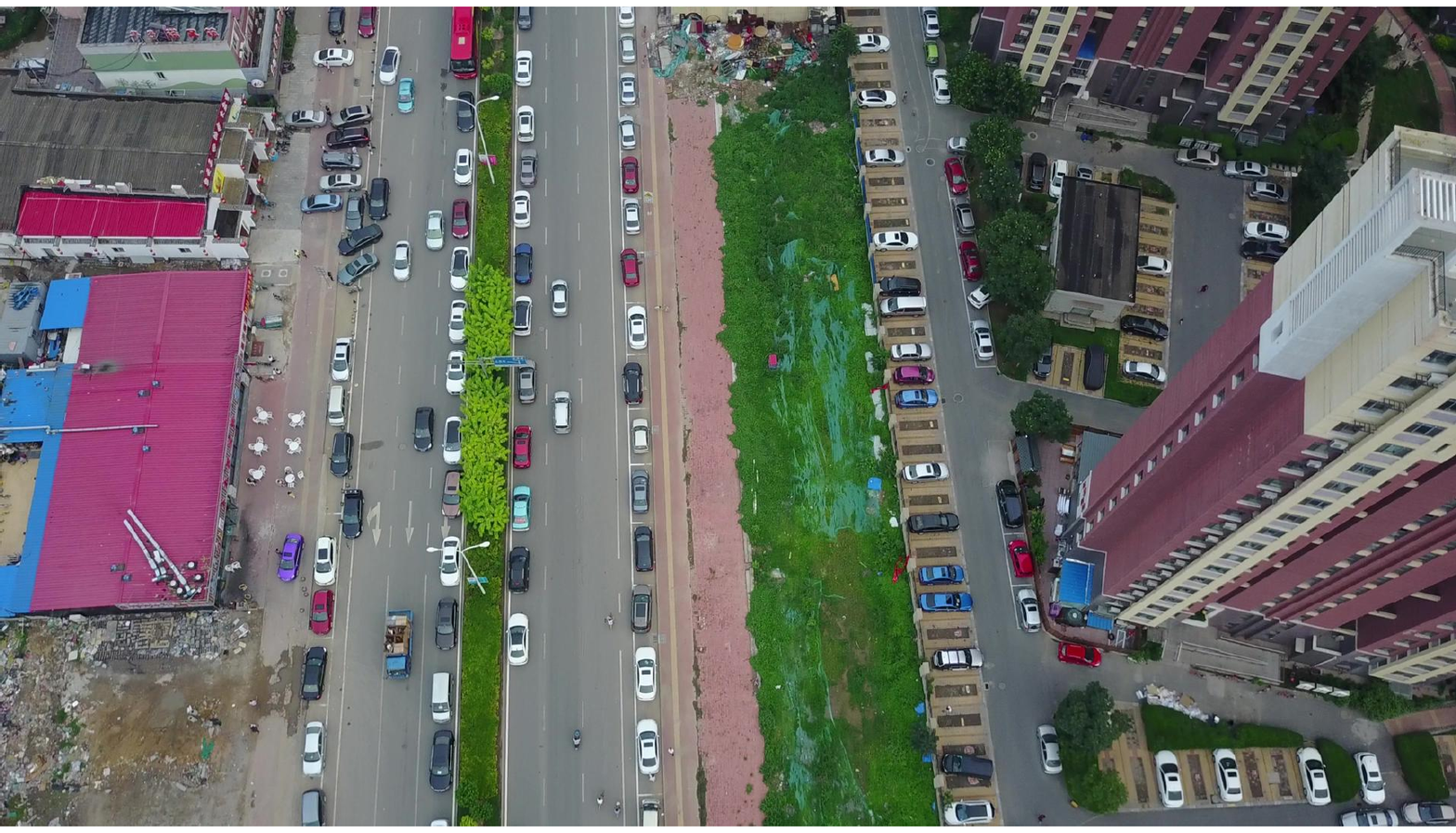}}
\end{minipage}
\tcblower
\blue{\textbf{Response :}} \\
The blue cars park on the right parking lot.\\
Coordinates: [0.628, 0.478], [0.646, 0.692], [0.649, 0.727] \\
The white cars park on both sides of the main road in a line.\\
Coordinates: [0.431, 0.051], [0.431, 0.15], [0.431, 0.101], [0.434, 0.256], [0.439, 0.523], [0.443, 0.813], [0.445, 0.902], [0.446, 0.982], [0.235, 0.43], [0.232, 0.486], [0.223, 0.674], [0.215, 0.904]

\end{tcolorbox}

\label{gpt-ass}
\end{table*}

\begin{table*}[t]\centering
\caption{System prompts employed at different stages of RDAnnotator, replacing the default system messages shown in Table~\ref{table9}}
\begin{minipage}{0.99\textwidth}\vspace{0mm}    \centering
\begin{tcolorbox} 
    \centering
    \small
     \hspace{-6mm}
    \begin{tabular}{p{0.99\textwidth}}
\label{system prompt}
\begin{minipage}{0.99\textwidth}\vspace{0mm}

\blue{\textbf{System prompt: Step 1 scene understanding}}\\ You are an AI visual assistant that specializes in providing clear and accurate descriptions of images without any ambiguity or uncertainty. Your descriptions should focus solely on the content of the image itself and avoid mentioning any location-specific details such as regions or countries where the image might have been captured.  \\ 

\blue{\textbf{System prompt: Step 2 color categorization}}\\As an AI visual assistant, your role involves analyzing a single image. \par
You are supplied with the specific attributes of objects within the image. This can include information about categories, colors, and precise coordinates. Such coordinates, represented as floating-point numbers that range from 0 to 1, are shared as center points, denoted as (x, y), identifying the center x and y.\par

Your task is to assess whether the given colors of specific objects match their appearance in the image. Respond with "Yes" when the colors are appropriate. In cases where the colors are deemed inappropriate, respond with a concise "No."\\

\blue{\textbf{System prompt: Step 3 expression generation}}\\As an AI visual assistant, your role involves analyzing a single image. You are supplied with three sentences that caption the image, along with additional data about specific attributes of objects within the image. This can include information about categories, colors, and precise coordinates. Such coordinates, represented as floating-point numbers that range from 0 to 1, are shared as center points, denoted as (x, y), identifying the center x and y. When coordinate x tends to 0, the object nears the left side of the image, shifting towards the right as coordinate x approaches 1. When coordinate y tends to 0, the object nears the top of the image, shifting towards the bottom as coordinate y approaches 1. \par
            
Your task is to classify the provided objects based on various characteristics, while also substantiating your classification. This classification should be thoroughly justified, with criteria including but not limited to relationships or relative locations of objects.\par

To refer to a specific object, use the provided coordinates directly. Base your classification justifications on direct observations from the image, avoiding any hypothesizing or assumptions.\\

\blue{\textbf{System prompt: Step 4 quality evaluation}}\\As an AI visual assistant, your role involves analyzing a single image. You are supplied with three sentences that describe the image, along with additional data about specific attributes of objects within the image. This can include information about categories, colors, and precise coordinates. Such coordinates, represented as floating-point numbers that range from 0 to 1, are shared as center points, denoted as (x, y), identifying the center x and y. When coordinate x tends to 0, the object nears the left side of the image, shifting towards the right as coordinate x approaches 1. When coordinate y tends to 0, the object nears the top of the image, shifting towards the bottom as coordinate y approaches 1. Besides, you are supplied with the description of the objects and their corresponding attributes.\par

Your task is to confirm whether the description exclusively relates to the described objects without including any others in the visual. Respond "yes" if it matches, or "no" with an explanation if it does not.
\end{minipage}
    \end{tabular}
\end{tcolorbox}
    
\vspace{-2mm}

\end{minipage}

\label{table11}
\end{table*}
\begin{table*}[t]\centering
\caption{RDAnnotator system prompts for feedback mechanism. Differences from Table~\ref{table11} are \textbf{highlighted}}
\begin{minipage}{0.99\textwidth}\vspace{0mm}    \centering
\begin{tcolorbox} 
    \centering
    \small
     \hspace{-6mm}
    \begin{tabular}{p{0.99\textwidth}}
\label{feedback system prompt}
\begin{minipage}{0.99\textwidth}\vspace{0mm}

\blue{\textbf{System prompt: Step 2 color categorization with feedback mechanism}}\\As an AI visual assistant, your role involves analyzing a single image. \par
You are supplied with the specific attributes of objects within the image. This can include information about categories, colors, and precise coordinates. Such coordinates, represented as floating-point numbers that range from 0 to 1, are shared as center points, denoted as (x, y), identifying the center x and y.\par

Your task is to assess whether the given colors of specific objects match their appearance in the image. Respond with "Yes" when the colors are appropriate. In cases where the colors are deemed inappropriate, respond with a concise "No."\\

\blue{\textbf{System prompt: Step 3 expression generation with feedback mechanism}}\\As an AI visual assistant, your role involves analyzing a single image. You are supplied with three sentences that caption the image, along with additional data about specific attributes of objects within the image. This can include information about categories, colors, and precise coordinates. Such coordinates, represented as floating-point numbers that range from 0 to 1, are shared as center points, denoted as (x, y), identifying the center x and y. When coordinate x tends to 0, the object nears the left side of the image, shifting towards the right as coordinate x approaches 1. When coordinate y tends to 0, the object nears the top of the image, shifting towards the bottom as coordinate y approaches 1. \textbf{You are also provided with descriptions and the objects that initially failed to match, along with the reasons for the discrepancies.}\par

\textbf{Your task is to revise both the description and the corresponding objects to correct these mismatches based on the provided reasons. Ensure that the revised description accurately matches the corresponding objects depicted in the visual content.}\\

\blue{\textbf{System prompt: Step 4 quality evaluation with feedback mechanism}}\\As an AI visual assistant, your role involves analyzing a single image. You are supplied with three sentences that describe the image, along with additional data about specific attributes of objects within the image. This can include information about categories, colors, and precise coordinates. Such coordinates, represented as floating-point numbers that range from 0 to 1, are shared as center points, denoted as (x, y), identifying the center x and y. When coordinate x tends to 0, the object nears the left side of the image, shifting towards the right as coordinate x approaches 1. When coordinate y tends to 0, the object nears the top of the image, shifting towards the bottom as coordinate y approaches 1. \textbf{Besides, you are supplied with the description and the objects that initially failed to match.}\par

\textbf{Your task is to provide detailed reasoning for unsuccessful object matches.
}

\end{minipage}
    \end{tabular}
\end{tcolorbox}

\end{minipage}

\label{table12}
\end{table*}

\end{document}